\documentclass[letterpaper, 10 pt, conference]{ieeeconf}  

\IEEEoverridecommandlockouts                              

\usepackage{graphics} 
\usepackage{epsfig} 
\usepackage{times} 
\usepackage{amsmath} 
\usepackage{amssymb}  
\usepackage{mathrsfs}
\usepackage[T1]{fontenc}
\usepackage[linesnumbered,ruled]{algorithm2e}
\usepackage{mathrsfs}
\usepackage[table]{xcolor}
\usepackage{subfigure} 
\usepackage{cite}

\usepackage{hyperref}
\hypersetup{
	colorlinks=true,
	linkcolor=blue,
	filecolor=blue,      
	urlcolor=blue,
	citecolor=cyan,
}
\usepackage{booktabs}
\title{\LARGE \bf
Training-Set Distillation for Real-Time UAV Object Tracking
}

\author{Fan Li$^{1}$, Changhong Fu$^{1,*}$, Fuling Lin$^{1}$, Yiming Li$^{1}$, and Peng Lu$^{2}$
	\thanks{$^{1}$Fan Li, Changhong Fu, Fuling Lin and Yiming Li are with the School of Mechanical Engineering, Tongji University, 201804 Shanghai, China.
		{\tt\small changhongfu@tongji.edu.cn}}%
	\thanks{$^{2}$Peng Lu is with the Adaptive Robotic Controls Lab (ArcLab), Hong Kong Polytechnic University (PolyU), Hong Kong, China.
		{\tt\small peng.lu@polyu.edu.hk}}%
	\thanks{The source code and UAV tracking videos are available in \url{https://github.com/vision4robotics/TSD-Tracker} and \url{https://youtu.be/2RYDYtqZFBA}. }
}

\begin{document}

\maketitle
\thispagestyle{empty}
\pagestyle{empty}

\begin{abstract}
Correlation filter (CF) has recently exhibited promising performance in visual object tracking for unmanned aerial vehicle (UAV). 
Such online learning method heavily depends on the quality of the training-set, yet complicated aerial scenarios like occlusion or out of view can reduce its reliability. 
In this work, a novel time slot-based distillation approach is proposed to efficiently and effectively optimize the training-set's quality on the fly. 
A cooperative energy minimization function is established to score the historical samples adaptively. 
To accelerate the scoring process, frames with high confident tracking results are employed as the keyframes to divide the tracking process into multiple time slots. 
After the establishment of a new slot, the weighted fusion of the previous samples generates one key-sample, in order to reduce the number of samples to be scored. 
Besides, when the current time slot exceeds the maximum frame number, which can be scored, the sample with the lowest score will be discarded. 
Consequently, the training-set can be efficiently and reliably distilled. 
Comprehensive tests on two well-known UAV benchmarks prove the effectiveness of our method with real-time speed on a single CPU.
\end{abstract}

\section{Introduction}
With the development of unmanned aerial vehicle (UAV), visual tracking plays an increasingly important role in prosperous practical applications, such as obstacle avoidance~\cite{Fu2014ICRA}, aerial refueling~\cite{Yin2016TIM}, autonomous landing~\cite{Lin2017AR}, \textit{etc}. 
Appreciable progress has been made in UAV tracking in recent years. 
However, it still faces many challenges, including strong UAV/object motion, frequent viewpoint change, severe illumination variation, abnormal appearance variation (occlusion or out-of-view). 
Additionally, mobile aerial platform has increased the difficulty in tracking scenarios such as mechanical vibration, restricted computational capability, limited power capacity, to name a few.

\begin{figure}[t]
	\centering
	\includegraphics[width=0.49\textwidth]{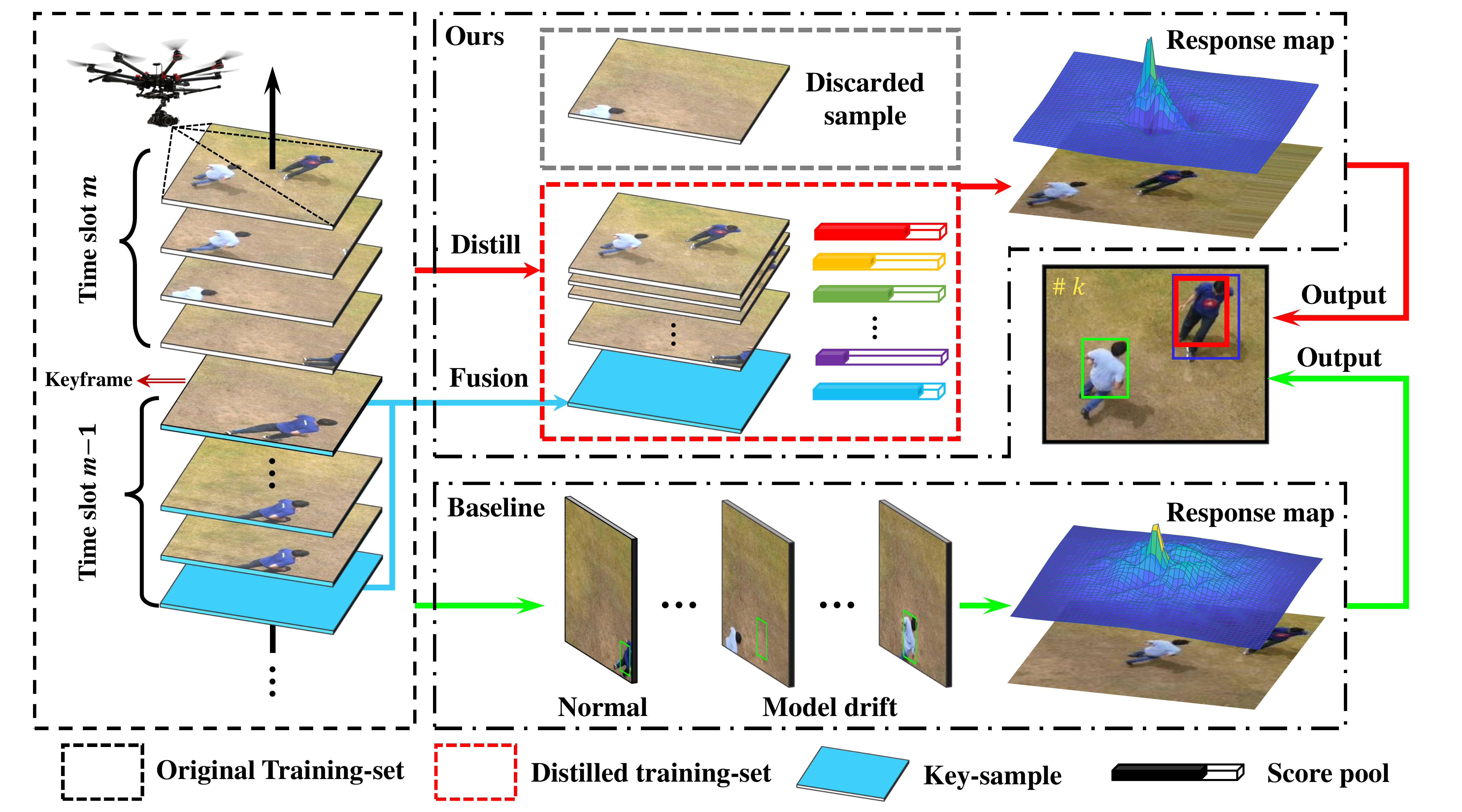}
	\caption{Comparison between our tracker with the baseline~\cite{Galoogahi2017ICCV}. Blue box is the ground truth. Red and green ones are the results of ours and baseline, respectively. The latest established time slot (time slot $m-1$) is fused into one key-sample. The frame number in the current time slot (time slot $m$) keeps increasing until the next keyframe arrives. It is noted that the distilled training-set has a capacity, indicating that the number of samples to be scored cannot be too large for efficiency reason. Therefore, when the current time slot possesses more frames than the distilled training-set capacity, the sample with the lowest score will be discarded. The discarding continues to be implemented until the next sample with high confident tracking result arrives, afterwards, current training-set will be fused into one key-sample again. In light of the adaptive scoring, the influence of unreliable samples can be repressed using our method, while the baseline suffers from drift.}
	\label{fig:start}
\end{figure}

In recent years,  correlation filter (CF)-based approaches \cite{Henriques2014TPAMI} have become a widely used framework for visual tracking due to its outstanding computational efficiency obtained in Fourier domain, especially for aerial tracking tasks in which the computational resources are precious. 
The training sample set used to learn the CF model is collected from the video frames. 
In traditional CF-based methods, to help CF adapt to the object appearance change, the online learned CF model is updated with new samples frame-by-frame via a simple linear interpolation method. 
Unfortunately, this strategy can be easily affected by unreliable samples, which can be introduced by many challenges, \textit{e.g.}, occlusion and out-of-view, leading to suboptimal performance, as shown in Fig.~\ref{fig:start}.

Existing CF-based trackers either ignore the problem mentioned above~\cite{LukezicCVPR2017,Huang2019ICCV,Zhang2017PR,Mueller2017CVPR,Fu2019RS} or directly adjust the learning rate for model update if the new samples are criticized as unreliable~\cite{Wang2017CVPR,Bolme2010CVPR,Zhang2017CVPR}. 
An efficacious approach is to manage training-set via explicit component~\cite{Gao2014ECCV,Zhang2014ECCV}. 
However, they have to process large prior samples in each frame, leading to an undesired increase in the computational burden. 

In the online tracking process, when a new frame arrives, CF-based tracker firstly obtains a response map by correlating the learned filter and new samples, and then the object location is predicted according to the peak in the map~\cite{Li2014ECCVws,Johnander2017CAIP,Danelljan2016ECCV}. 
The quality of the response map, such as the sharpness and the fluctuation, can to some extent reflect the confidence degree about the tracking result \cite{Huang2019ICCV}. 
In this work, the response map is exploited to measure the reliability of training samples. 
Frames with high confident tracking results are used as keyframes to divide the tracking into multiple time slots. 
Besides, the response map is also integrated into a joint correlation filter and sample weight optimization framework, which is developed to score the training sample adaptively. 
To speed up the scoring process, the training-set will be fused into one key-sample once the new time slot is established. 
Besides, if the capacity of training-set is larger than the scoring capacity, the samples with the lowest score will be discarded. 
Our core contributions are: 

\begin{itemize}
	\item A novel \textbf{t}ime \textbf{s}lot-based \textbf{d}istillation algorithm for UAV tracking (TSD) is proposed to enhance the quality of the training-set efficiently. 
	\item Extensive and comprehensive tests on 193 challenging UAV image sequences have demonstrated that TSD tracker achieves competitive performance against the state-of-the-art works and runs at real-time frame rates.
\end{itemize}

\section{Related works}\label{sec:RELATEDWORK}
Discriminative correlation filter-based tracker was firstly introduced in the minimum output sum of squared error (MOSSE)~\cite{Bolme2010CVPR} filer. 
J. F. Henriques \textit{et al.}~\cite{Henriques2014TPAMI} further introduced the kernel trick into CF-based framework. 
Computing efficiency was improved greatly by employing the circulant matrix property and solving the regression problem in the Fourier domain. 
By introducing modern multi-dimensional features, the accuracy of CF-based approaches is further improved~\cite{Danelljan2014BMVC,Danelljan2014CVPR,Fan2017ICCV}. 
However, as the CF-based trackers usually obtain negative samples by shifting images cyclically, the trained filter is influenced undesirably by boundary effect. 
Existing approaches~\cite{Danelljan2015ICCV, Li2018CVPR, Danelljan2017CVPR} have made up for this shortcoming by spatial regularization. 
To utilize real background information, learning background-aware correlation filters (BACF)~\cite{Galoogahi2017ICCV} enlarged search regions to extract real negative training examples from the background. 
However, those methods are susceptible to unreliable samples which were introduced by occlusion, out of view, viewpoint change and other reasons.

\begin{figure}[t]
	\centering
	\includegraphics[width=0.47\textwidth]{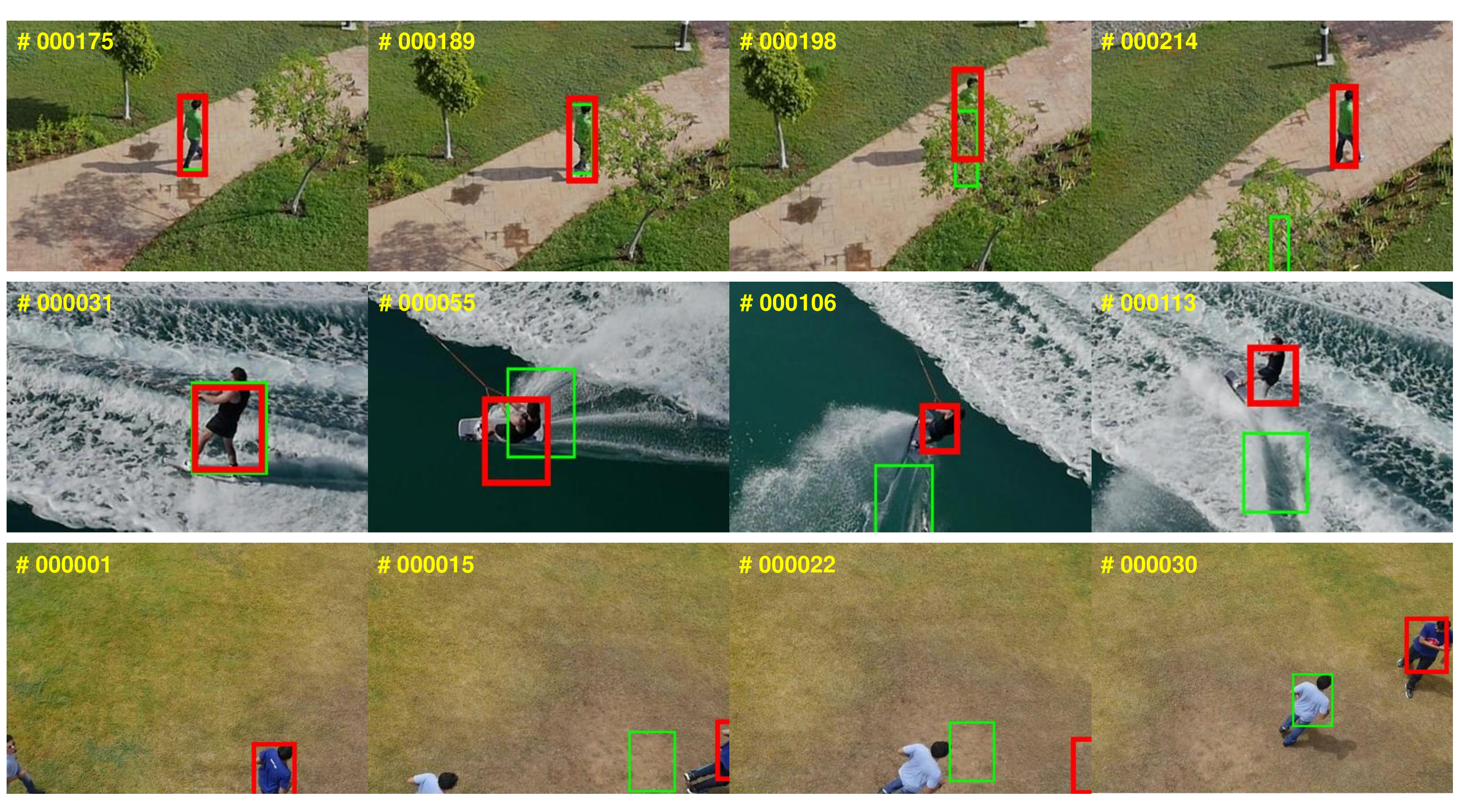}
	\caption{A comparison of the proposed TSD tracker (red) with the baseline BACF~\cite{Galoogahi2017ICCV} (green) in  $person17\_1$, $wakeboard5$ and $person9$. In all the three sequences, BACF suffers from handling unreliable samples, leading to drift problem in cases of object occlusion (top row), viewpoint change (middle row), and out of view (bottom row). TSD tracker successfully employs reliable samples in filter training, generating more robust   model.}
	\label{fig:purification_main}
	\vspace{-13pt}
\end{figure}

Most existing methods choose not to update the model if the samples do not meet certain criteria. 
D. S. Bolme \textit{et al.}~\cite{Bolme2010CVPR} and M. Wang \textit{et al.}~\cite{Wang2017CVPR} rejected new samples based on the peak-to-sidelobe ratio (PSR) and peak-to-correlation energy (APCE), respectively. 
However, due to some challenging scenarios like viewpoint change, some samples that were previously considered reliable may turn into unreliable samples, as shown in the middle row of Fig.~\ref{fig:purification_main}. 
Those samples may lead to model drift. 
Other methods exploit explicit component to manage training-set. 
In \cite{Gao2014ECCV}, an explicit component based on distance comparisons is used to manage the training-set, and J. Zhang \textit{et al.}~\cite{Zhang2014ECCV} used a combination of experts to correct undesirable model updates. 
A unified formulation for discriminative tracking~\cite{Danelljan2016CVPR} evaluated the quality of the samples to manage the training-set dynamically. 
This type of method can improve the quality of the training-set efficaciously. 
However, they have to process large samples, thereby struggling to meet real-time performance requirements for UAV tracking scenarios.

\section{Proposed tracking approach}\label{sec:METHOD}
\begin{table}[t]
	\scriptsize
	\setlength{\tabcolsep}{1mm}
	\centering
	\caption{A brief description of the variables in this work.}
	\begin{tabular}{cc}
		\hline\hline
		\textbf{Symbol}&\textbf{Description} \\\hline
		$\mathbf{x}_{[d]}^k\in\mathbb{R}^{N}$&the $d$-th channel of the sample from $k$-th frame\\
		$k$&the sequence number of the current frame\\
		$\mathbf{x}^{key}_{[d],m}\in\mathbb{R}^{N}$&the $d$-th channel of key-sample fused in the $m$-th time slot\\
		$k_m$&the sequence number of the $m$-th keyframe\\
		$\mathbf{x}^f_{[d]}\in\mathbb{R}^{N}$&the $d$-th channel of the $f$-th sample in the training-set\\
		$f$& the frame number in current time slot, \textit{i.e.}, $f= k-k_m+1$\\
		$\mathbf{y}_j\in\mathbb{R}^{N}$&the $j$-th element of the predetermined ideal response\\
		$\mathbf{w}_{[d]}\in\mathbb{R}^{M}$&the $d$-th channel of the learned correlation filter\\
		${\alpha}^f\in \mathbb{C}$&the score of the $f$-th sample in the training-set\\
		\hline\hline
	\end{tabular}%
	\label{Table:key_components}%
	\vspace{-5pt}
\end{table}%
In this section, the baseline BACF is firstly reviewed for better understanding, then our TSD tracker is presented. The main symbols in this work are summarized in Table \ref{Table:key_components}.

\subsection{Revisiting BACF}
Given the $d$-th channel of the vectorized training samples $\mathbf{x}_{[d]}$ and predetermined vectorized ideal response $\mathbf{y}_{[d]} \in \mathbb{R}^{N}$, the objective of training the filter $\mathbf{w}$ is to minimize:
\begin{equation}
\small
\begin{split}
\mathcal{E}(\mathbf{w})
=
\sum_{j=1}^{N}\left\|\mathbf{y}_j
-
\sum_{d=1}^{D}\mathbf{w}_{[d]}^{\top}\mathbf{B}\mathbf{x}_{[d]}[\Delta\boldsymbol{\tau}_j]\right\|_{2}^{2}
+
\frac{\lambda}{2}\sum_{d=1}^{D}\left\|\mathbf{w}_{[d]}\right\|_{2}^{2}
\end{split}
\ ,
\end{equation}
\noindent where $[\Delta\boldsymbol{\tau}_j]$ is the circular shift operator, and $\mathbf{x}_{[d]}[\Delta\boldsymbol{\tau}_j]$ denotes 
$j$-step circular shifted sample $\mathbf{x}_{[d]}$. 
$\mathbf{w}_{[d]} \in \mathbb{R}^{M}$ denotes the learned correlation filter.
$\mathbf{B}$ is a $M \times N$ binary matrix which crops the mid $D$ elements of sample $\mathbf{x}_{[d]}$. It is worth noting that $N \gg M$. 

Though BACF has achieved satisfactory performance owing to the augment of real-world negative samples, it still adopts a simple linear interpolation method to update the training samples, raising the possibility of introducing unreliable information like occlusion or outdated appearance. 
In this work, a time slot-based joint filter and sample score optimization framework is proposed built on BACF. 
Specifically, the training-set is restricted to a small size for raising the optimization speed. 
When the number of current training samples exceeds the given size, the most unreliable sample will be discarded. 
In addition, frames with high confident tracking results are employed as the keyframes to divide the tracking process into multiple time slots. 
Once the slot is established, the samples are fused into one key-sample with their weights.
\subsection{Overall objective of TSD}
In this work, a novel dynamic time slot-based distillation approach is constructed. 
The proposed filter $\mathbf{w}$ and sample scores $\alpha$ can be collaboratively learned by minimizing a cooperative energy minimization function as follows:
\begin{equation}
\small
\begin{split}
\mathcal{E}(\mathbf{w},\mathbf{\alpha})
=
\mathcal{E}_1(\mathbf{w},\mathbf{\alpha})
+
\mathcal{E}_2(\mathbf{\alpha})
+
\mathcal{E}_3(\mathbf{\alpha})
\end{split}
\ ,
\label{Eq:zhugongshi}
\end{equation}

\noindent where each term is described below.
\\
\subsubsection{\textbf{Classification error $\mathcal{E}_1$}} Similar to BACF, given a training-set $\{\mathbf{x}^1,\mathbf{x}^2,...,\mathbf{x}^F\}$, the loss for the discrepancy between the scheduled response and the filter response for the sample $\mathbf{x}^f$ is defined as:
\begin{equation}
\small
\begin{split}
\mathcal{E}_1(\mathbf{w},\mathbf{\alpha})
=&
\sum_{f=1}^{F}\left(\alpha^{f}\sum_{j=1}^{N}\left\|\mathbf{y}_j
-
\sum_{d=1}^{D}\mathbf{w}_{[d]}^{\top}\mathbf{B}\mathbf{x}_{[d]}^{f}[\Delta\boldsymbol{\tau}_j^{f}]\right\|_2^2\right)\\
&+
\frac{\lambda}{2}\sum_{d=1}^{D}\left\|\mathbf{w}_{[d]}\right\|_2^2
\end{split}
\ ,
\label{Eq:E1}
\end{equation}

\noindent where $\mathbf{x}^{f}$ is the $f$-th sample in the training-set. 
$\alpha^{f}$ is the score of sample $\mathbf{x}^f$. 
$\lambda$ is a regularization parameter.

Different from BACF which sets the score $\alpha^{f}\in\mathbb{R}$ using a fixed learning rate parameter, in TSD, different samples have distinct scores based on their confidence. 
Reliable samples have higher scores, and vice versa. The number of training samples $F$ is defined by:
\begin{equation}
\small
\begin{aligned}
F = min\{f, F_{max}\}
\end{aligned}
\ ,
\label{Eq:K_max}
\end{equation}
\noindent where $f$ denotes the frame number in the current time slot. To avoid exceeding the upper limit of memory consumption, the training-set capacity $F_{max}$ is proposed. 
If the number of samples exceeds $F_{max}$, the sample with the lowest score will be discarded.

\subsubsection{\textbf{Temporal regularization $\mathcal{E}_2$}} To account for fast appearance changes, recent samples are given larger scores:
\begin{equation}
\small
\begin{split}
\mathcal{E}_2(\alpha)
=
\frac{\gamma}{2}\sum_{f=1}^{F}\frac{(\alpha^{f})^2}{t^{f}}
\end{split}
\ ,
\label{Eq:e2}
\end{equation}
\noindent where $\gamma$ is a trade-off parameter between classification error and temporal regulation. 
$t^f$ is a function related to the distance between the current frame and the last selected keyframe for time slot division. It is designed as follows:
\begin{equation}
\small
t^f=\begin{cases}
a^{-1},
\quad\quad\quad\quad\quad\quad\quad f=1,...,F-f_0\\
a^{-1}(1-q)^{F-f_0-f},
\quad f=F-f_0+1,...,F
\end{cases}
\ ,
\label{Eq:temporal}
\end{equation}
\noindent where the constant $a=F-f_0+\frac{(1-q)^{-f_0}-1}{q}$  is determined by the condition $\sum_{f=1}^{F}t^{f}=1$. 
In this work, $f_0$ and $q$ are set to 10 and 0.0408, respectively.\\

\subsubsection{\textbf{Response map-based regularization $\mathcal{E}_3$}} The confidence degree about the tracking result can be reflected by the quality of the response map to some extent. 
Thus the samples with higher quality responses are given higher scores:
\begin{equation}
\small
\begin{split}
\mathcal{E}_3(\alpha)
=
\frac{\nu}{2}\sum_{f=1}^{F}
\frac{(\alpha^{f})^2}{{DPMR}^f}
\end{split}
\ .
\label{Eq:e3}
\end{equation}
\noindent where $\nu$ is a trade-off parameter between classification error and response map-based regulation. 
In this work, the quality of the response map is evaluated by the proposed dual-area peak to media ratio ($DPMR$) that is described is in Section~\ref{section:DPMR}.

\subsection{Optimization algorithm}\label{Optimization}

To unify the dimensions of each term in Eq.~(\ref{Eq:zhugongshi}), intermediate variable $h_{[d]}$ is introduced as follows:
	\begin{equation}
	\small
	\begin{split}
	\mathbf{h}_{[d]}
	=
	\begin{bmatrix} 0,\mathbf{w}_{[d]}^{\top},0 \end{bmatrix}^{\top}
	\end{split}
	\ .
	\end{equation}
	
Thus Eq.~(\ref{Eq:zhugongshi}) can be expressed in the frequency domain as:
	\begin{equation}
	\small
	\begin{split}
	\mathcal{E}
	=&
	\sum_{f=1}^{F}\left(\alpha^{f}\left\|{\hat y}_{u,[d]}
	-
	\sum_{d=1}^{D}{\hat{g}}_{u,[d]}^*{\hat{x}}_{u,[d]}^{f}\right\|_2^2\right)
	+
	\frac{\lambda}{2}\left\|{\hat{h}}_{u,[d]}^*\right\|_2^2\\
	&+
	\frac{\gamma}{2}\sum_{f=1}^{F}\frac{(\alpha^{f})^2}{t^{f}}
	+
	\frac{\nu}{2}\sum_{f=1}^{F-1}DPMR^f(\alpha^{f})^2\\
	&\ \ \ \ \ \textit{s.t.}\ \ 
	\hat{g}_{u,[d]}^* =\hat{h}_{u,[d]}^*\\
	\end{split}
	\ ,
	\label{Eq:frequent}
	\end{equation}
\noindent where $\hat{\ }$ denotes the Discrete Fourier Transform (DFT) of a signal, and${\ }^{*}$ denotes complex conjugate. 
As for the subscript $u$, it represents the element in a data set (\textit{i.e.}, $\hat{x}_u$ refers to the element in a training-set). 
Different from BACF, all the operations in Eq.~(\ref{Eq:frequent}) are performed in element-wise.

To learn parameter $\mathbf{\hat{w}^*}$ and $\alpha$, \textit{i.e.}, for learning parameter $\mathbf{\hat{h}^*}$ and $\alpha$, the problem is how to optimize Eq.~(\ref{Eq:frequent}). 
Two components of parameter can be optimized iteratively, \textit{i.e.}, parameter $\mathbf{\hat{h}^*}$ trained with fixed $\alpha$ firstly and then the process is inverted. 

\subsubsection{\textbf{Subproblem $\hat{\mathbf{h}}^{*}$}}

$\hat{\mathbf{h}}^{*}$ can be solved using Augmented Lagrangian Method (ALM):
\begin{equation}
\small
\label{Eq:ALM}
\begin{split}
\mathcal{L}
=&
\sum_{f=1}^{F}\left(\alpha^{f}\left\|{\hat{y}}_{u,[d]}
-
{\hat{g}}_{u,[d]}^*{\hat{x}}_{u,[d]}^{f}\right\|_2^2\right)
+
\frac{\lambda}{2}\left\|{\hat{h}}_{u,[d]}^*\right\|_2^2\\
+&
\hat{{\zeta}}_u^{\top}\left({\hat{g}}_{u,[d]}^*
-
{\hat{h}}_{u,[d]}^*\right)
+
\frac{\mu}{2}\left\|
{\hat{g}}_{u,[d]}^*
-
{\hat{h}}_{u,[d]}^*
\right\|_2^2\\
\end{split}
\ ,
\end{equation}

\noindent where $\mu$ and $\hat{\zeta}_u \in \mathbb{R}$ denote the penalty factor and the element of the Lagrangian vector in the Fourier domain separately. 
With the ADMM~\cite{Stephen2010FTML} technique, Eq.~(\ref{Eq:ALM}) can be solved as follows:
\begin{equation}
\small
\begin{split}
{\hat{h}}_{u,[d]}^*
=&
\arg \min _{{\hat{h}}_{u,[d]}}\Bigg\{
\frac{\lambda}{2}\left\|{\hat{h}}_{u,[d]}^*\right\|_2^2\\
&+
\hat{{\zeta}}^{\top}\left({\hat{{g}}}_{u,[d]}^*
-
{\hat{{h}}}_{u,[d]}^*\right)
+
\frac{\mu}{2}\left\|
{\hat{g}}_{u,[d]}^*
-
{\hat{h}}_{u,[d]}^*
\right\|_2^2\Bigg\}\\
\hat{g}_{u,[d]}^*
=&
\arg \min _{\hat{g}_{u,[d]}^{F}}\Bigg\{
\sum_{f=1}^{F}\left(\alpha^{f}\big\Vert\hat{y}_u
-
\hat{g}_{u,[d]}^*\hat{x}_{u,[d]}^{f}\big\Vert_2^2\right)\\
&+
\hat{{\zeta}}^{\top}\left(\hat{{g}}_{u,[d]}
-
\hat{{h}}_{u,[d]}\right)
+
\frac{\mu}{2}\big\Vert
\hat{{g}}_{u,[d]}
-
\hat{{h}}_{u,[d]}
\big\Vert_2^2\Bigg\}\\
\end{split}
\ .
\end{equation}

Thanks to the element-wise operation, the solution to both subproblems $\mathbf{\hat{g}}^*$ and $\mathbf{h}^*$ can be easily obtained as follows:
	\begin{equation}
	\small
	\begin{split}
	{\hat{h}}_{u,[d]}^*
	=&
	\left(\lambda+\mu\right)^{-1}\left(2\hat{{\zeta}}^{\top}+\mu{\hat{g}}_{u,[d]}\right)\\
	\hat{g}_{u,[d]}^*
	=&
	\left(
	\sum_{f=1}^{F}\alpha^{f}(\hat x_{u,[d]}^f)^*\hat x_{u,[d]}^f+\frac{\mu}{2}
	\right)^{-1}\\
	&\ \ \ \ \ \left(
	\sum_{f=1}^{F}\alpha^{f}\hat{y}_{u}^*\hat{x}_{u,[d]}^{f}
	-
	\hat{\zeta}^{\top}
	+
	\frac{\mu}{2}(\hat{h}_{u,[d]}^{F})^*
	\right)\\
	\end{split}
	\ .
	\end{equation}

The Lagrangian parameter in Eq.~(\ref{Eq:ALM}) is updated as follows:
\begin{equation}
\small
\begin{split}
\label{Eq:update}
\hat{\zeta}^{(j+1)}_{u,[d]}=\hat{\zeta}^{(j)}_{u,[d]}+\mu\left(\hat{\mathrm{g}}^{(j+1)*}_{u,[d]}-\hat{\mathrm{h}}^{(j+1)*}_{u,[d]}\right)
\ ,
\end{split}
\end{equation}
\noindent where superscript $(j)$ denotes the initial value or the value in the last iteration, and
subscript $(j+1)$ denotes the value at current iteration. 
Note that the subscript $u$ refers to the iterations of variable, and $\mathbf{h}_{[d]}=[0,\mathbf{w}_{[d]}^{\top},0]^{\top}$.

\subsubsection{\textbf{Subproblem $\hat{\mathbf{\alpha}}$}}
The second step of optimization is to train $\alpha$ with fixed $\hat{\mathbf{h}}^{*}$, which requires setting the first derivative of $\alpha$ to zero. Therefore, $\alpha$ can be obtained by:
	\begin{equation}
	\small
	\begin{aligned}
	\frac{\partial \mathcal{\hat{J}}}{\partial \mathcal{\alpha}^f}
	=&
	\frac{\partial}{\partial \mathcal{\alpha}^f}\Bigg(
	\sum_{f=1}^{F}\bigg(\alpha^{f}\Big\|\hat{y}_{u,[d]}
	-
	\sum_{d=1}^{D}(\hat{g}_{u,[d]}^{F})^*\hat{x}_{u,[d]}^{f}\Big\|_2^2\bigg)\\
	&+
	\frac{\gamma}{2}\sum_{f=1}^{F}\frac{(\alpha^{f})^2}{t^{f}}
	+
	\frac{\nu}{2}\sum_{f=1}^{F}DPMR^f(\alpha^f)^2\Bigg)\\
	\end{aligned}
	\ ,
	\end{equation}
\begin{figure}[t]
	\centering
	\includegraphics[width=0.48\textwidth]{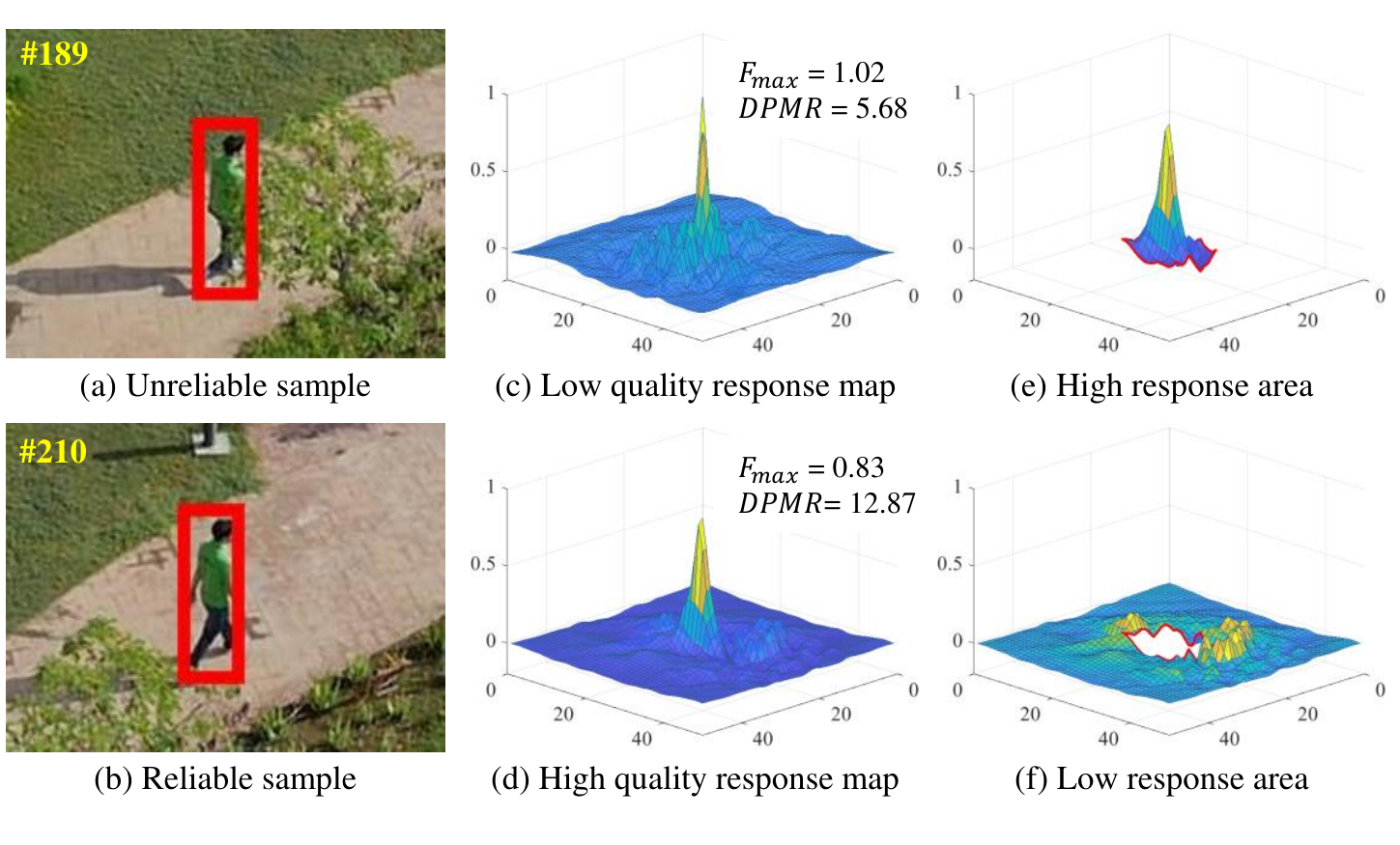}
	\caption{Visualization of the calculation of $DPMR$. The first column are images from example sequence $person17\_1$. The response maps in the second column come from TSD. The third column are the high and the low response area split from the response maps in (d).}
	\label{fig:PSRandPPSR}
	\vspace{-5pt}
\end{figure}

\noindent the above subproblem is equivalent to the quadratic programming problem as follows:
	\begin{equation}
\small
\begin{aligned}
\mathrm{min}\ \ \ \ \ \hat{\mathcal{J}}=&\sum_{f=1}^{F}(\beta^{f}\alpha^f+\gamma^{f}(\alpha^f)^{2})\\
\ \ \ \ \ \ \ \textit{s.t.}\ &\ \sum_{f=1}^{F}\alpha^f=1\\
\end{aligned}
\ .
\end{equation}

This optimization problem is a convex quadratic programming method. 
Therefore, $\alpha$ can be solved efficiently via standard quadratic programming.

\subsection{Time slot establishment}\label{section:DPMR} 
The quality of the response map can indicate the reliability of the tracking result. 
As illustrated in $(b)$ and $(d)$ of Fig.~\ref{fig:PSRandPPSR}, the response map with only one sharp peak is of high quality, and it should be smooth in the other areas. 
As for the tracking result with a low confidence degree, the response map will fluctuate intensely as shown in $(a)$ and $(c)$ of Fig.~\ref{fig:PSRandPPSR}. 
Traditional CF-based trackers use the peak value of the response map to evaluate its quality, which is inaccurate in many cases as shown in Fig.~\ref{fig:PSRandPPSR}. 
In this work, the quality of the response map is evaluated by the dual-area peak to media ratio ($DPMR$).

To calculated the $DPMR$, the response map is split into the high response area and the low response area as shown in $(e)$ and $(f)$ of Fig.~\ref{fig:PSRandPPSR}. 
The $DPMR$ is then defined as:

\begin{equation}
\small
\begin{aligned}
DPMR=\frac{\max(R_h)-\min(R_h)}{\operatorname{mean}\left(R_l\right)-\min(R_l)}
\end{aligned}
\ ,
\end{equation}
\noindent where $R_h$ and $R_l$ denote the high and low response area, respectively. 
$DPMR$ can indicate the fluctuated level of response map, which can reflect the confidence degree of the detected target. Finally, our strategy is designed as:

\begin{figure*}[!t]
	\centering
	\subfigure{
		\begin{minipage}[t]{0.48\textwidth}
			\includegraphics[width=1\textwidth]{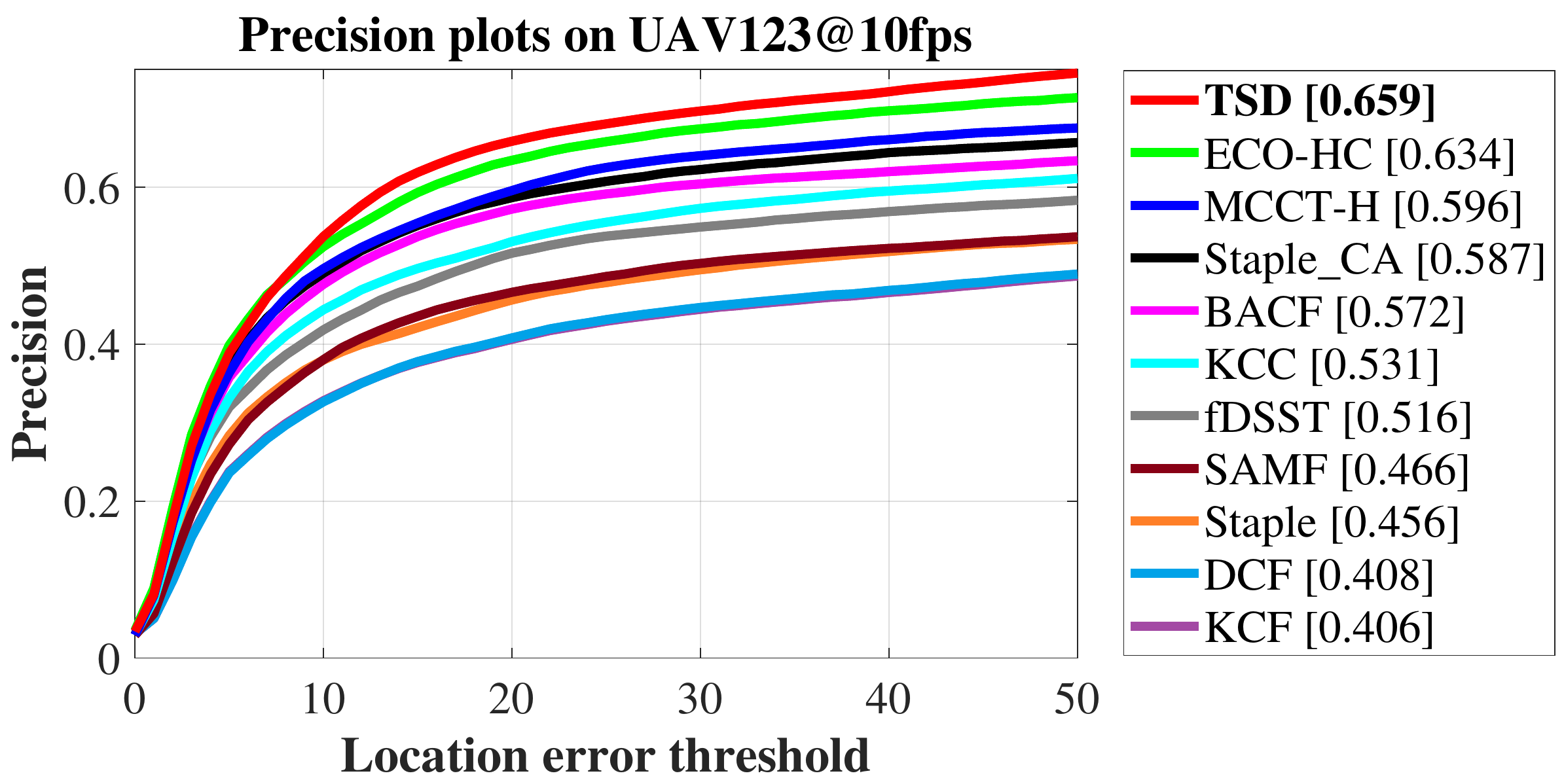}
			\includegraphics[width=1\textwidth]{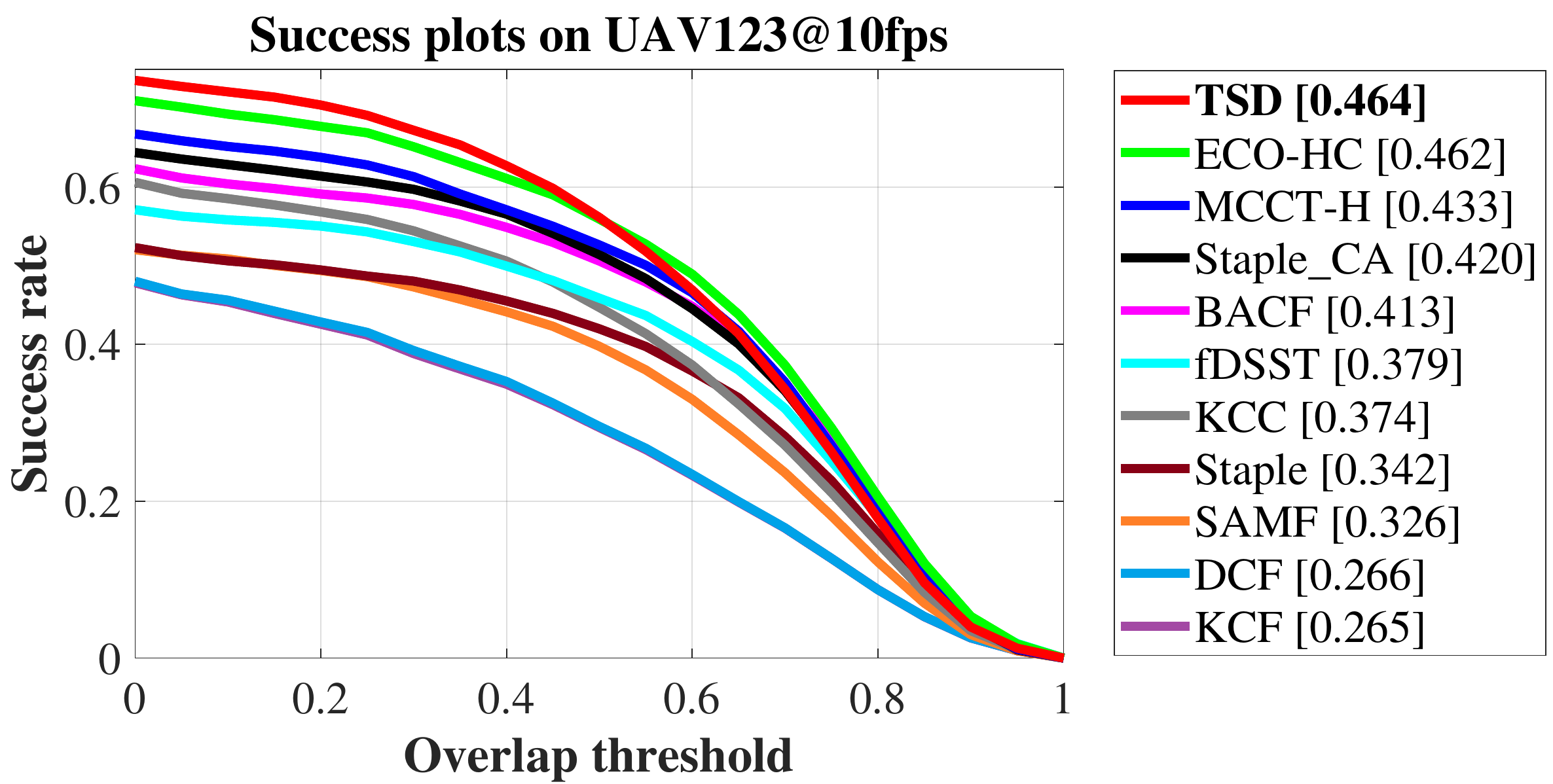}
		\end{minipage}%
		\label{fig:UAV123@10fps_performance}
	}%
\subfigure{
\begin{minipage}{0.48\textwidth}
	\includegraphics[width=1\textwidth]{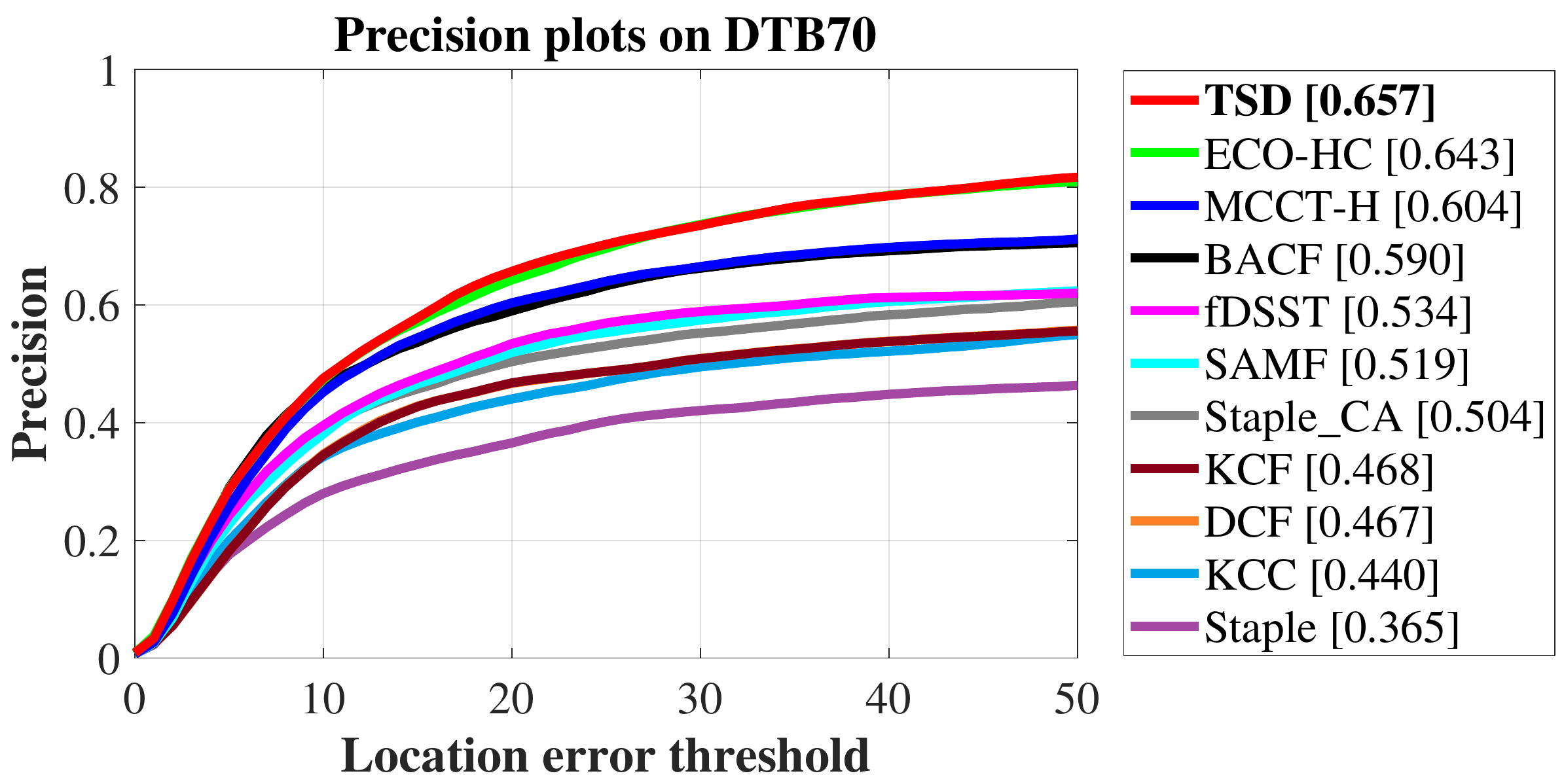}
	\includegraphics[width=1\textwidth]{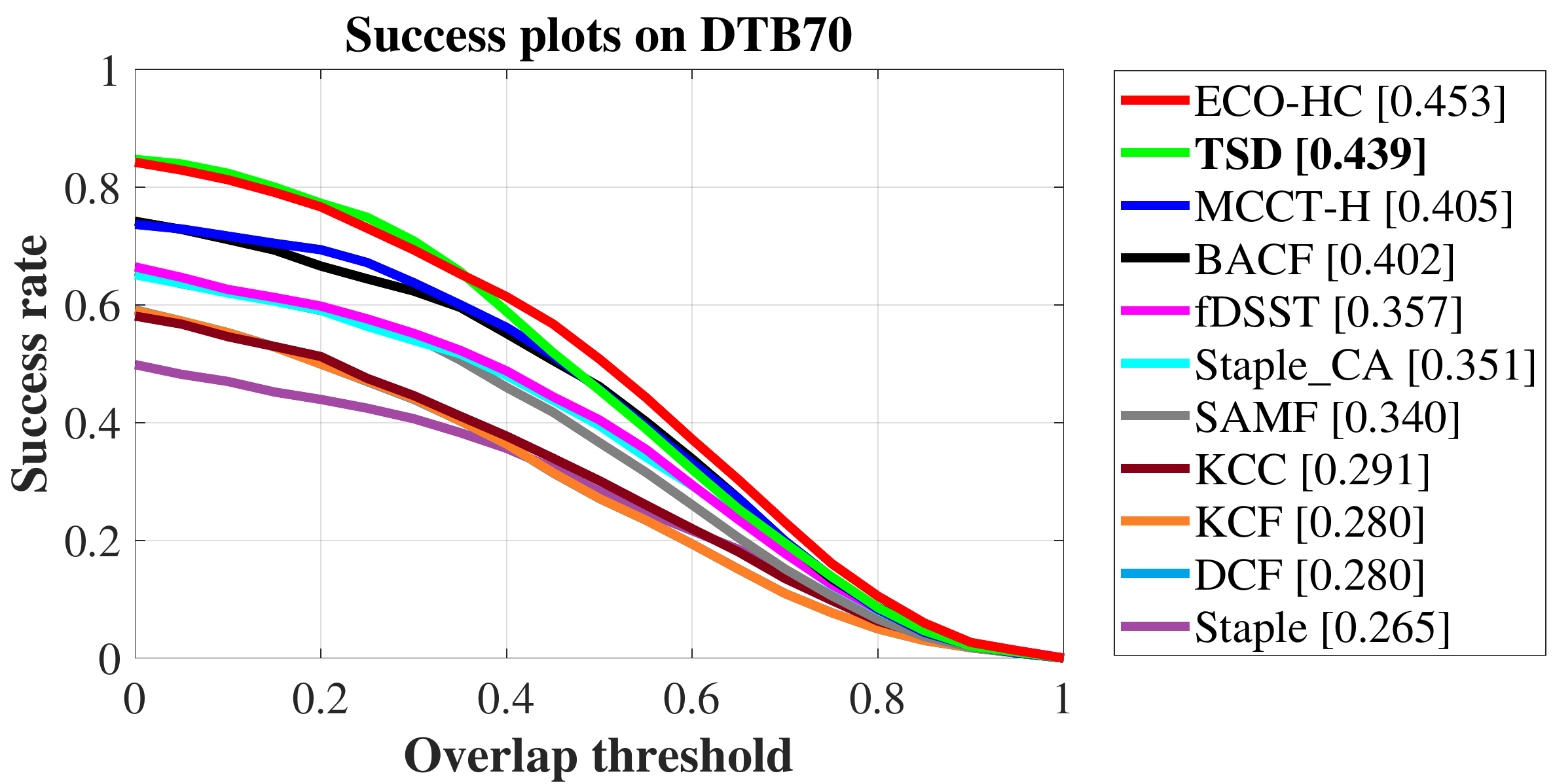}
\end{minipage}%
\label{fig:DTB70_performance}
}%
	\centering
	\caption{Precision and success plots of TSD as well as other state-of-the-art real-time trackers on a single CPU. Two standard evaluation measures are employed in precision and success plots, \textit{i.e.}, center location error (CLE) and success rate based on one-pass evaluation (OPE) \cite{Wu2015TPAMI}. The CLE is defined as the Euclidean distance between the center of the estimated bounding box and ground-truth location, which can measure the precision. The success rate is characterized as the intersection over union (IoU) of the tracker bounding box and ground-truth bounding box, which can indicate the precision of scale estimation.}
	\label{fig:zongtu}
\end{figure*} 
\begin{table*}[t]
	\scriptsize
	\setlength{\tabcolsep}{1.8mm}
	\centering
	\caption{Frame per second (FPS) and millisecond per frame (MSPF) of real-time trackers using single CPU reported on UAV123@10fps. {\color{red}Red} , {\color{green}green}, and {\color{blue}blue} fonts indicate the first, second and third place, respectively.}
	\begin{tabular}{cccccccccccc}
		\hline\hline
		\textbf{Algorithms}&\textbf{TSD}&\textbf{DSST}\cite{Danelljan2014BMVC}&\textbf{BACF}\cite{Galoogahi2017ICCV}&\textbf{Staple}\cite{Bertinetto2016CVPR}&\textbf{Staple\_CA}\cite{Mueller2017CVPR}&\textbf{MCCT-H}\cite{Wang2018CVPR}&\textbf{ECO-HC}\cite{Danelljan2017CVPR}&\textbf{fDSST}\cite{Danelljan2017TPAMI}&\textbf{KCC}\cite{Wang2018AAAI}&\textbf{DCF}\cite{Henriques2014TPAMI}&\textbf{KCF}\cite{Henriques2014TPAMI} \\\hline
		\textbf{FPS}&41.89&72.71&46.51&62.48&56.46&59.01&62.19&\textcolor[rgb]{ 0,  0,  1}{132}&40.65&\textcolor[rgb]{ 1,  0,  0}{660.73}&\textcolor[rgb]{ 0,  1,  0}{337.53}\\
		\textbf{MSPF}&23.87&13.75&21.50&16.01&17.71&16.95&16.08&\textcolor[rgb]{ 0,  0,  1}{7.58}&24.60&\textcolor[rgb]{ 1,  0,  0}{1.51}&\textcolor[rgb]{ 0,  1,  0}{2.65}\\\hline\hline		
	\end{tabular}%
	\label{table:table1}%
\end{table*}%
\begin{equation}
\small
\begin{aligned}
\textit { breakpoint }(M)
=
\left\{\begin{array}{l}{1, \ DPMR>t r} \\ {0, \text { otherwise }}\end{array}\right.
\end{aligned}
\ .
\end{equation}

In practice, we choose $tr$ = 14 based on our empirical results. 
If the output of \textit { breakpoint } is 1, this frame will be treated as keyframe to divide the tracking process, and a new time slot is established. 
Then this slot is fused to one key-sample.


\subsection{Weighted fusion}
As mentioned above, the sample score $\alpha$ can evaluate the quality of each sample. 
To decrease the number of samples to be scored, at the end of a period of time slot, the last distilled training-set is fused into one key-sample $\mathbf{x}^{key}$ as follows:
\begin{equation}
\small
\begin{aligned}
\mathbf{x}^{key}=\sum_{f=1}^{F}\left({\alpha}^{f}\mathbf{x}^f\right)\ ,
\end{aligned}
\end{equation}
\noindent where $\mathbf{x}^{f}$ is the $f$-th sample in the last distilled training-set. 
$\alpha^f$ is the sample score of the sample in f-th selected frame. 
Key-sample is taken as the first sample of the next time slot.

\section{Experiments}\label{sec:experiments}
In this section, the proposed TSD tracker is evaluated  on 193 challenging UAV image sequences from two well-known and frequently-used benchmarks, \textit{i.e.}, UAV123@10fps~\cite{Mueller2016ECCV} and DTB70~\cite{LiAAAI2017}. 
The experimental results are compared to state-of-the-art trackers with real-time speed on a single CPU ($>$30fps), \textit{i.e.}, 
DSST~\cite{Danelljan2014BMVC}, 
BACF~\cite{Galoogahi2017ICCV}, 
Staple~\cite{Bertinetto2016CVPR}, 
Staple\_CA~\cite{Mueller2017CVPR}, 
MCCT-H~\cite{Wang2018CVPR}, 
ECO-HC~\cite{Danelljan2017CVPR}, 
fDSST~\cite{Danelljan2017TPAMI}, 
DCF~\cite{Henriques2014TPAMI}, KCF~\cite{Henriques2014TPAMI}, and 
KCC~\cite{Wang2018AAAI}.

\subsection{Implementation details}
TSD tracker is implemented in Matlab R2018a. 
Color names (CN) \cite{Danelljan2014CVPR} is adopted as the feature representation to raise efficiency. 
The training-set capacity $F_{max}$ in Eq.~(\ref{Eq:K_max}) is set to 50. 
$\gamma$ in Eq.~(\ref{Eq:e2}) and $\nu$ in Eq.~(\ref{Eq:e3}) are set to 3.02 and 0.201. In Eq.~(\ref{Eq:temporal}), $f_0$ = 10 and q = 0.0408. 
For ADMM iteration, $\mathbf{g}$, $\mathbf{w}$ and $\mathbf{h}$ are all initialized using null matrices, and update scale is set to 2. All the experiments are run on the computer with an i7-8700K (3.70KHz) CPU, 32GB RAM, and a single NVIDIA RTX 2080 GPU for fair comparisons. 
\begin{figure}[t]
	\centering
	\includegraphics[width=0.48\textwidth]{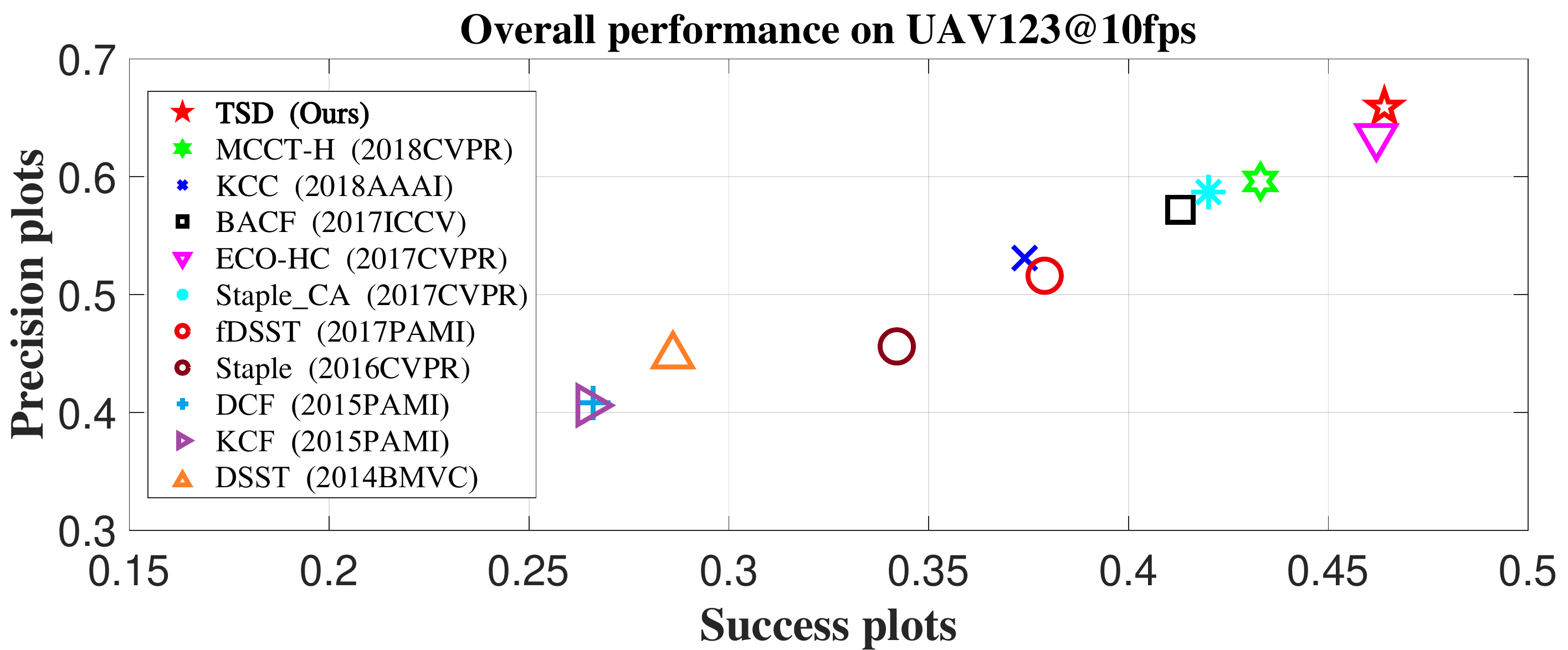}
	\caption{Ranking map of the real-time state-of-the-art trackers on CPU.}
	\label{fig:paiming}
	\vspace{-5pt}
\end{figure}
%


\subsection{Comparison with real-time trackers on a single CPU}
Due to the extremely harsh computational resources onboard UAVs, an energy-efficient and low-cost CPU is desirable in UAV tracking applications. In this work, we evaluate the trackers with real-time frame rates on a single CPU.
\begin{figure*}[!t]
	\centering
	\subfigure{
		\begin{minipage}[t]{0.32\textwidth}
			\centering
			\includegraphics[width=1\textwidth]{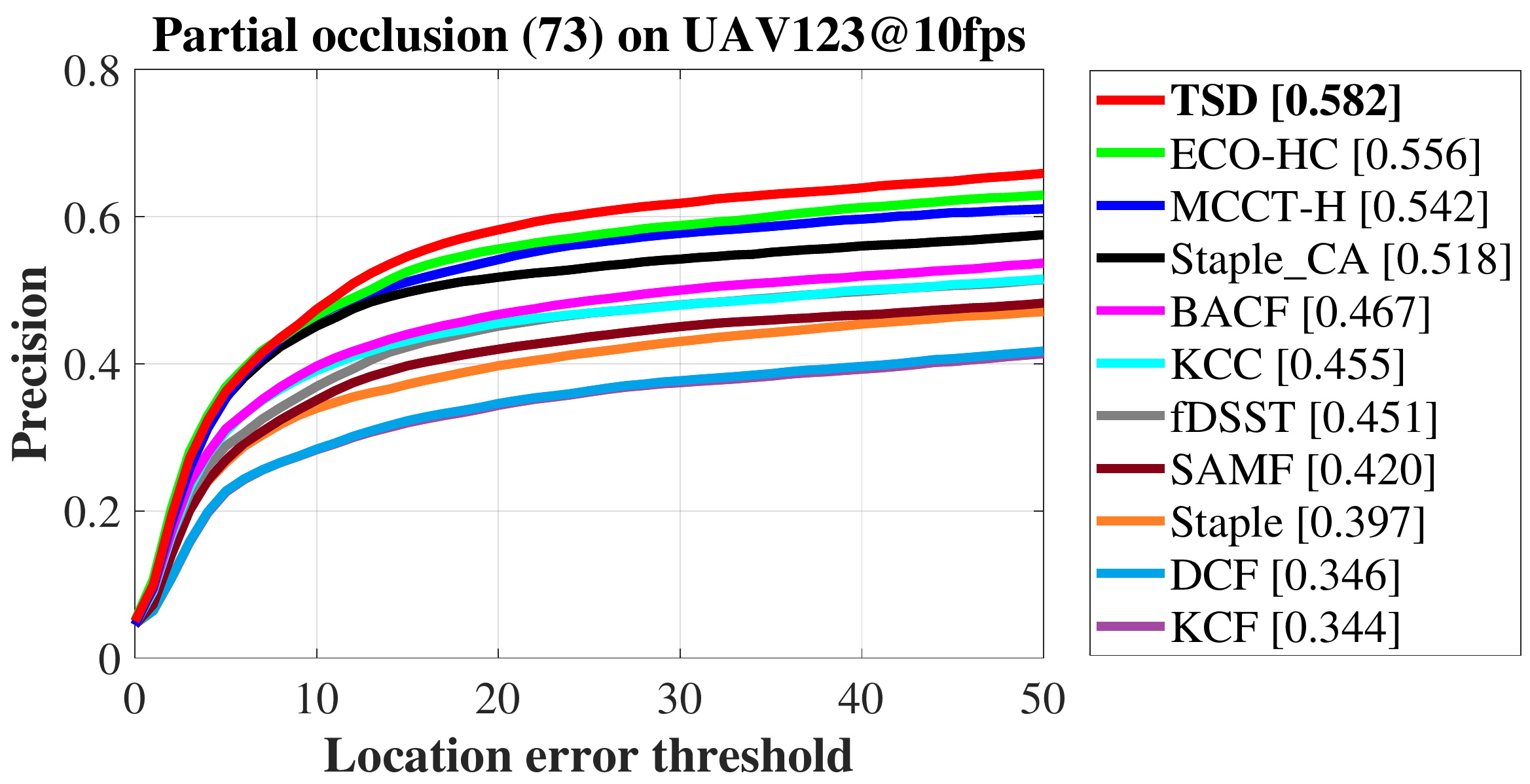}
		\end{minipage}%
	}%
	\subfigure{
		\begin{minipage}[t]{0.32\textwidth}
			\centering
			\includegraphics[width=1\textwidth]{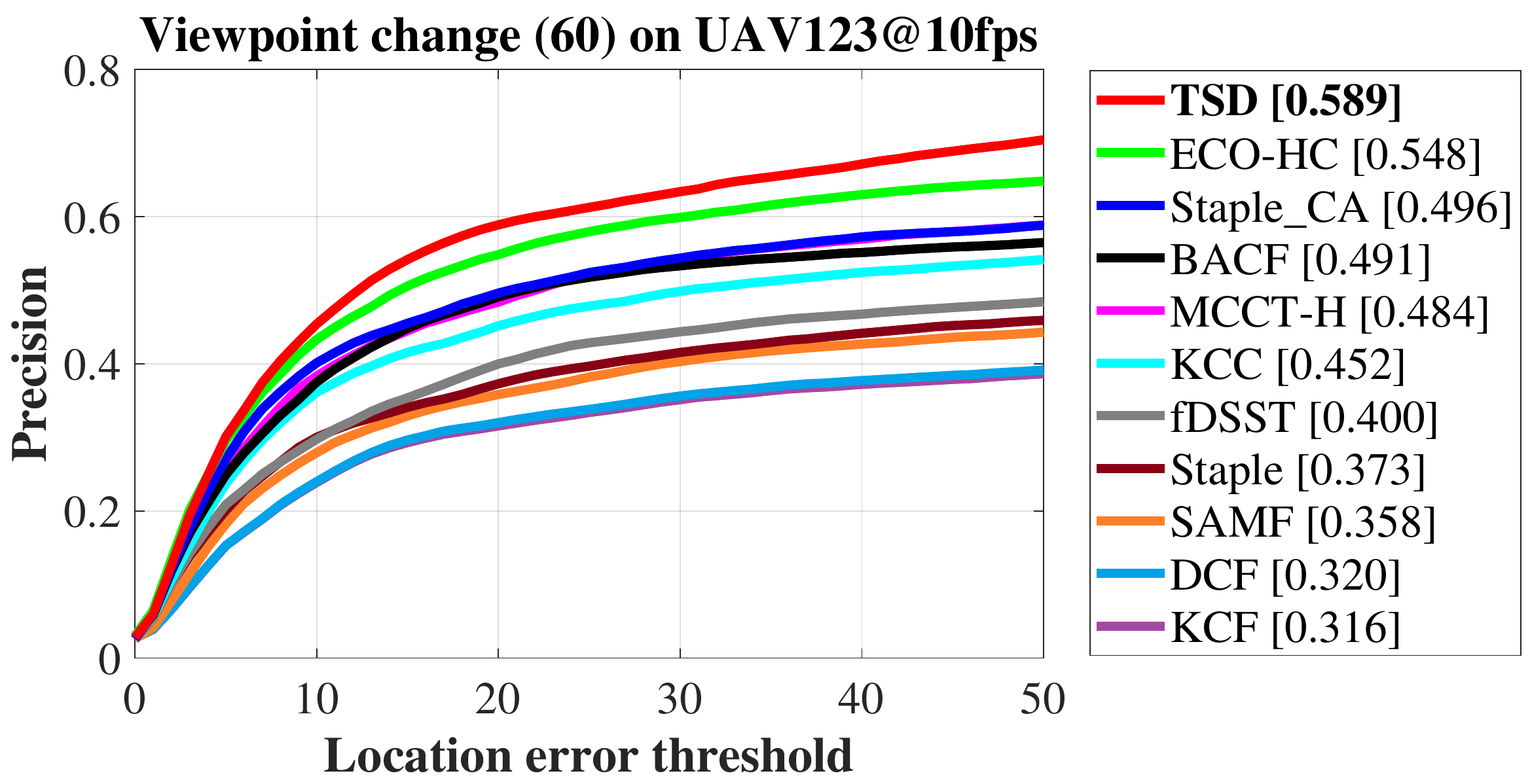}
		\end{minipage}%
	}%
	\subfigure{
		\begin{minipage}[t]{0.32\textwidth}
			\centering
			\includegraphics[width=1\textwidth]{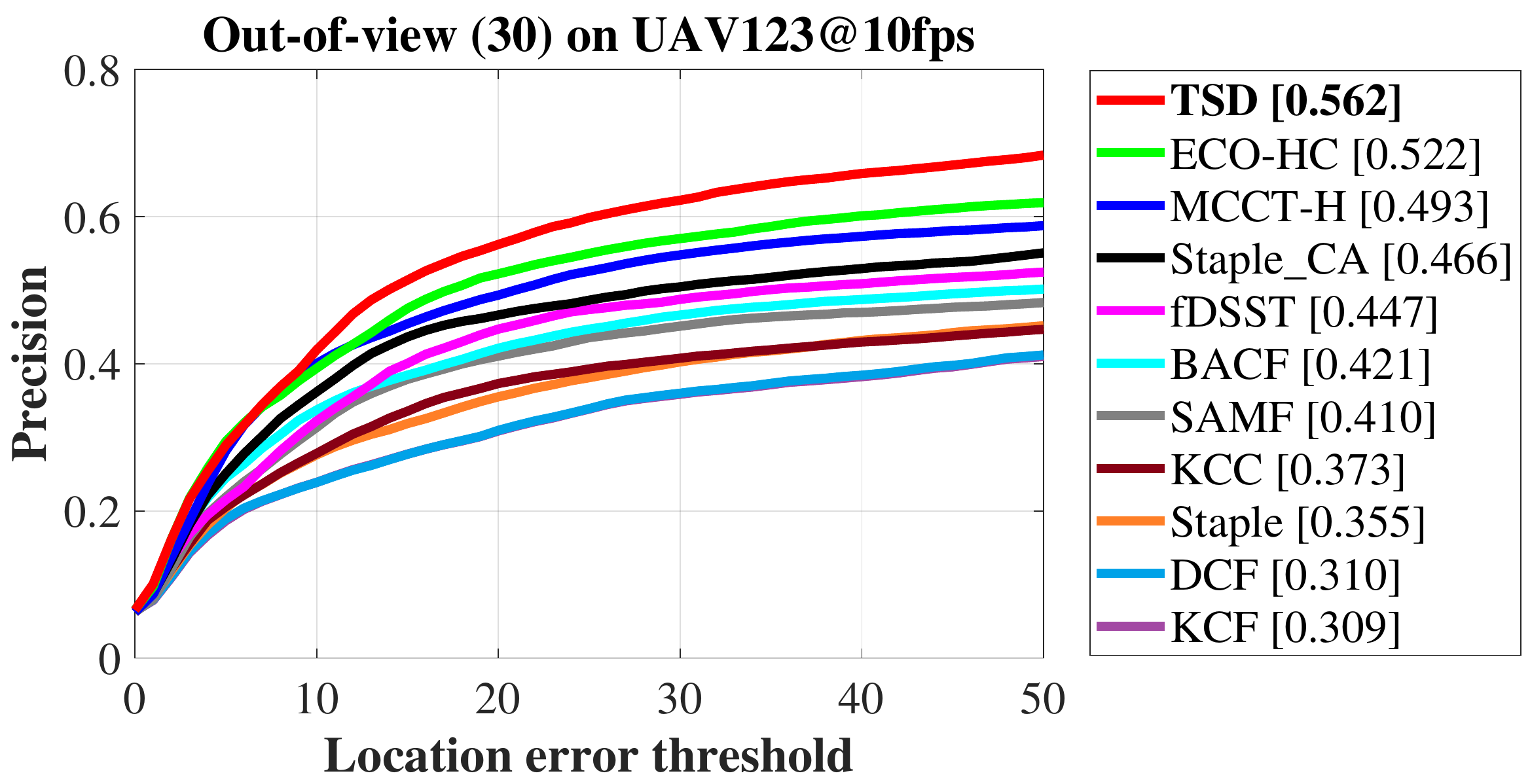}
		\end{minipage}%
	}%
	
	\subfigure{
		\begin{minipage}[t]{0.32\textwidth}
			\centering
			\includegraphics[width=1\textwidth]{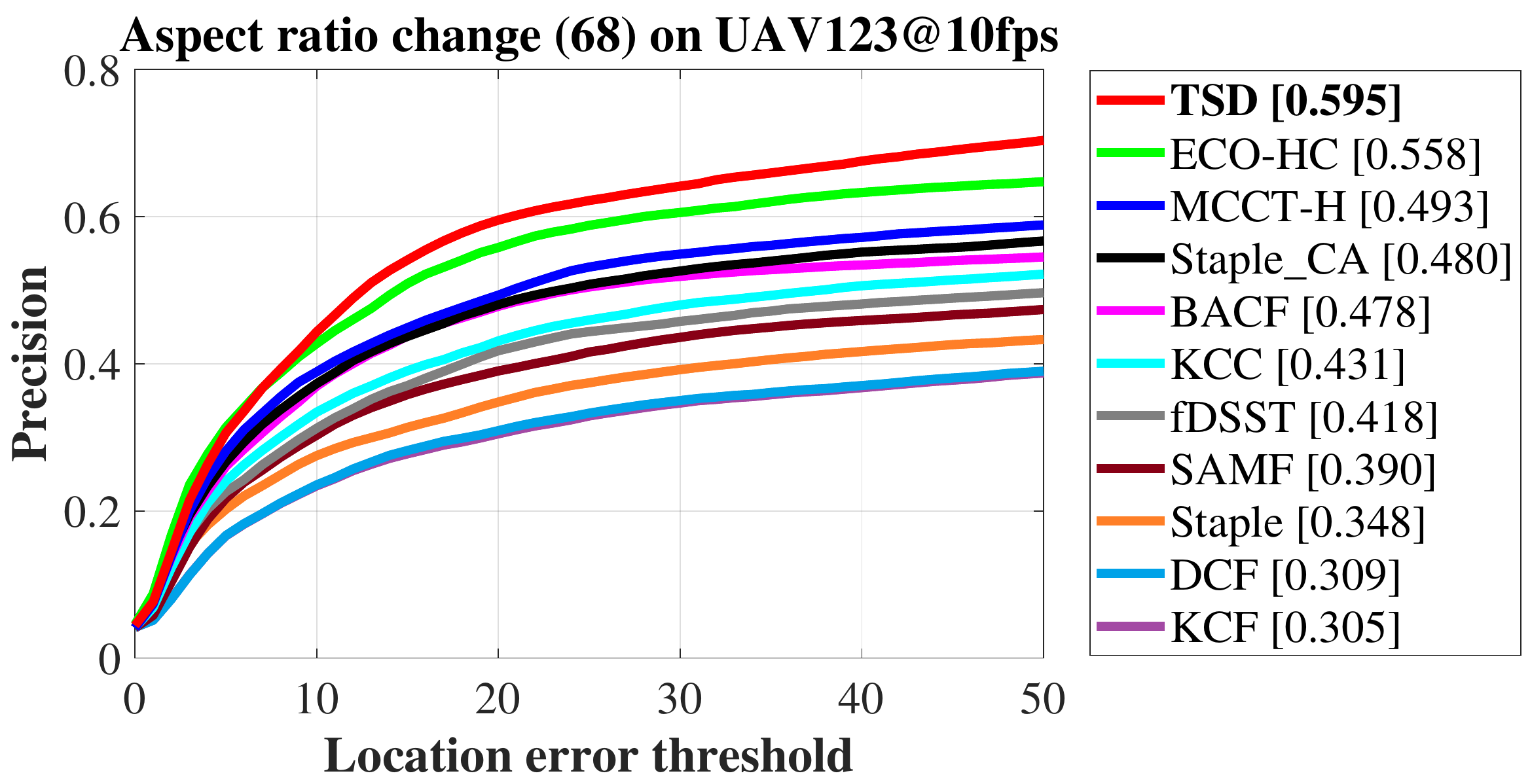}
		\end{minipage}%
	}%
	\subfigure{
		\begin{minipage}[t]{0.32\textwidth}
			\centering
			\includegraphics[width=1\textwidth]{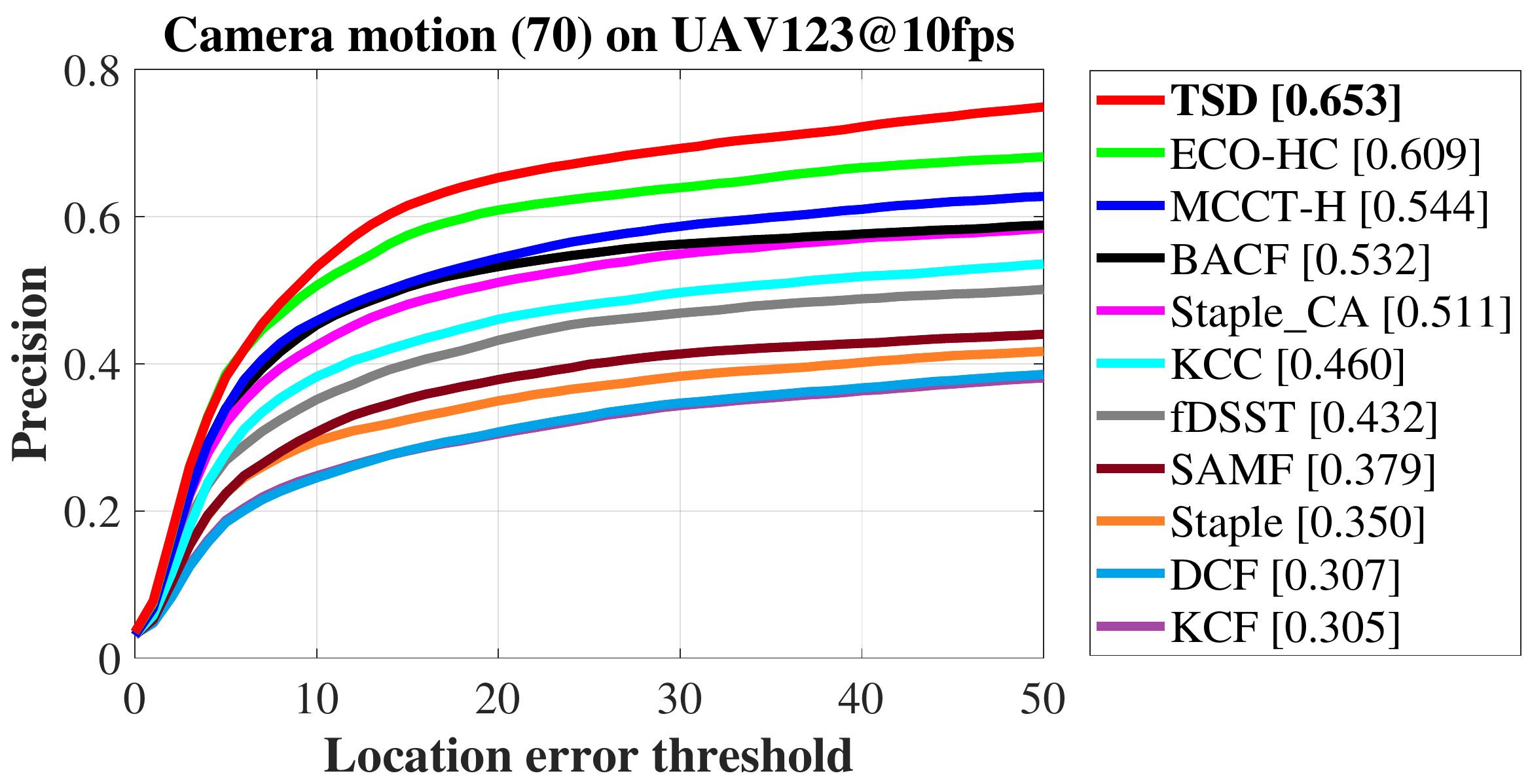}
		\end{minipage}%
	}%
	\subfigure{
		\begin{minipage}[t]{0.32\textwidth}
			\centering
			\includegraphics[width=1\textwidth]{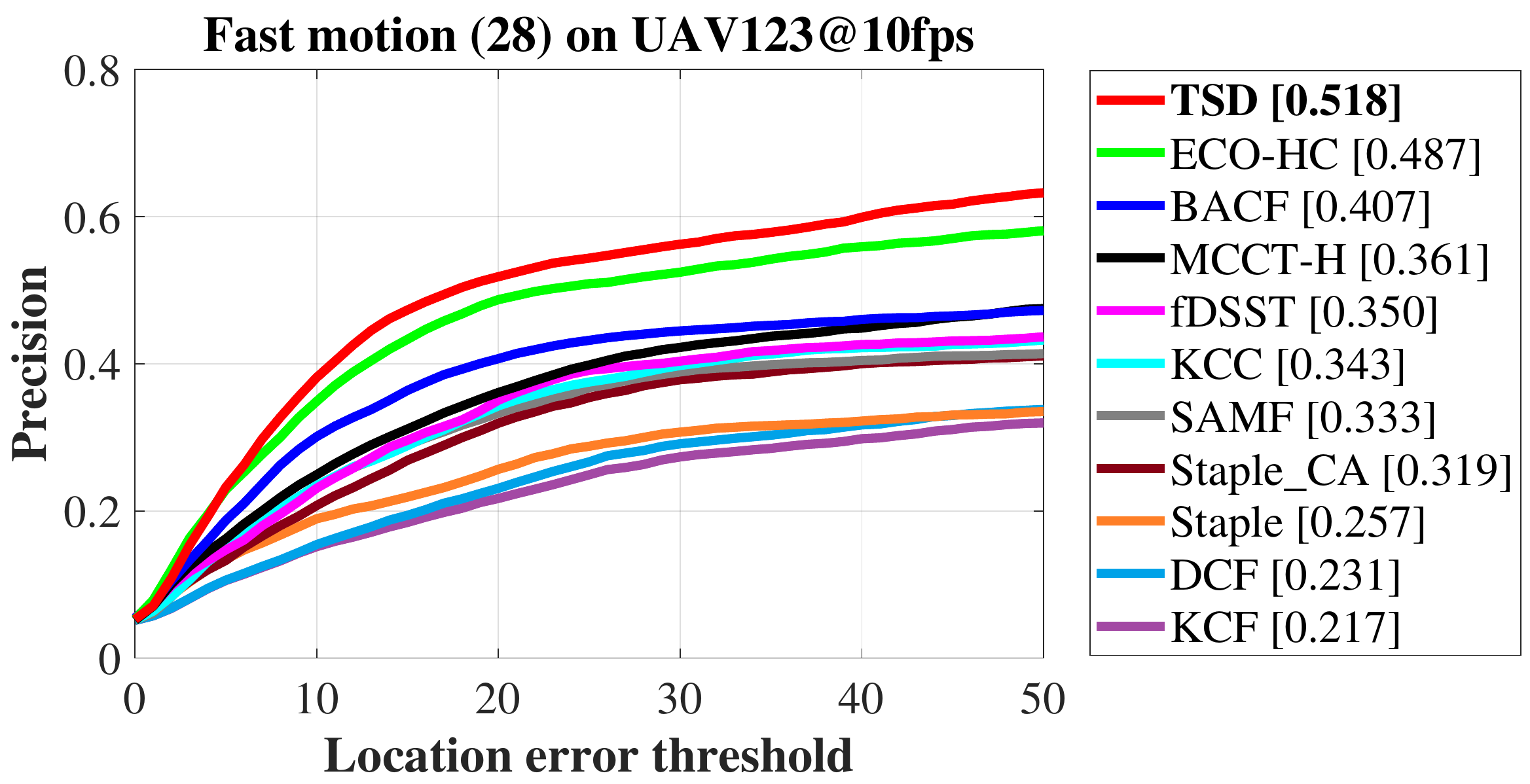}
		\end{minipage}%
	}%
	
	\subfigure{
		\begin{minipage}[t]{0.32\textwidth}
			\centering
			\includegraphics[width=1\textwidth]{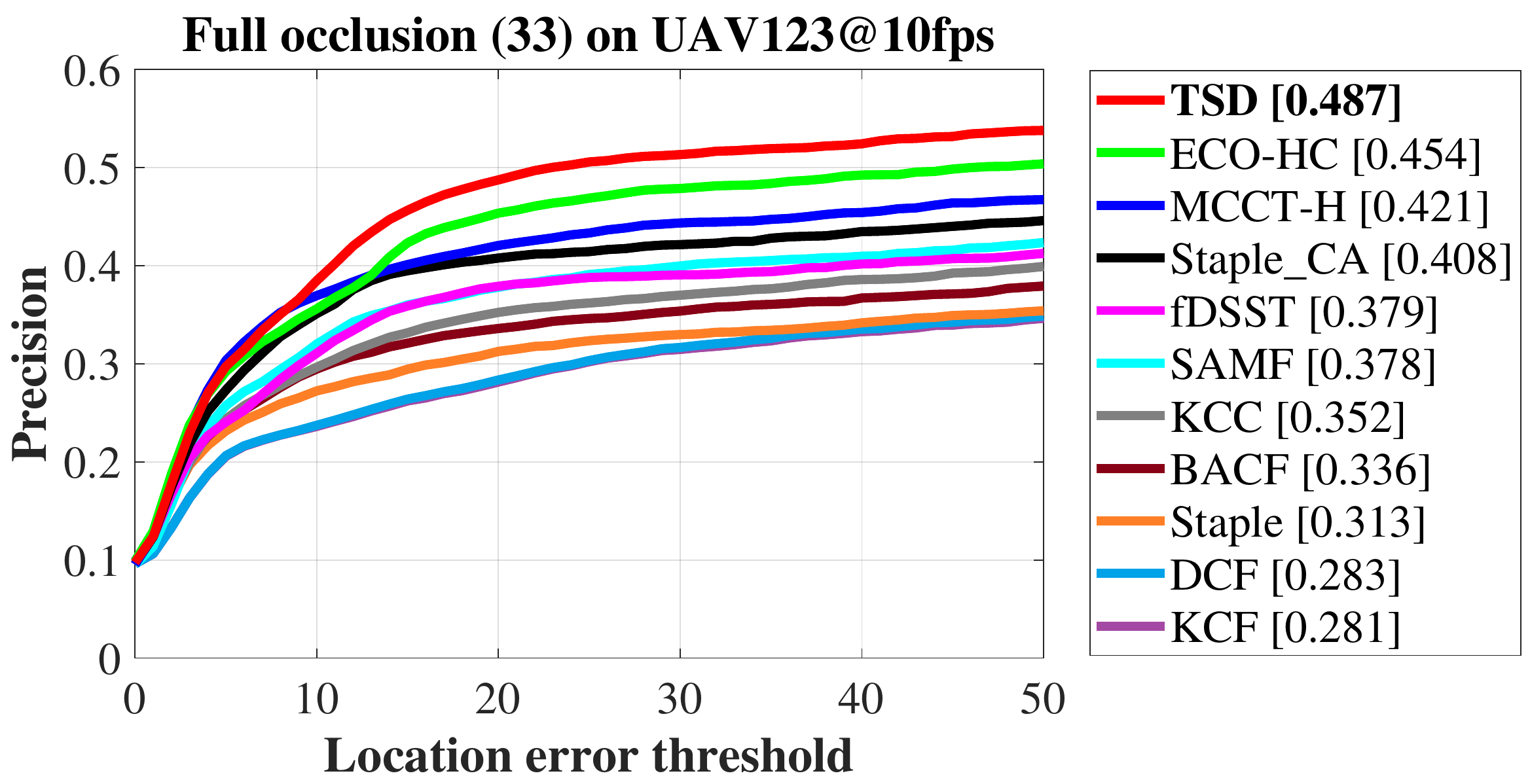}
		\end{minipage}%
	}%
	\subfigure{
		\begin{minipage}[t]{0.315\textwidth}
			\centering
			\includegraphics[width=1\textwidth]{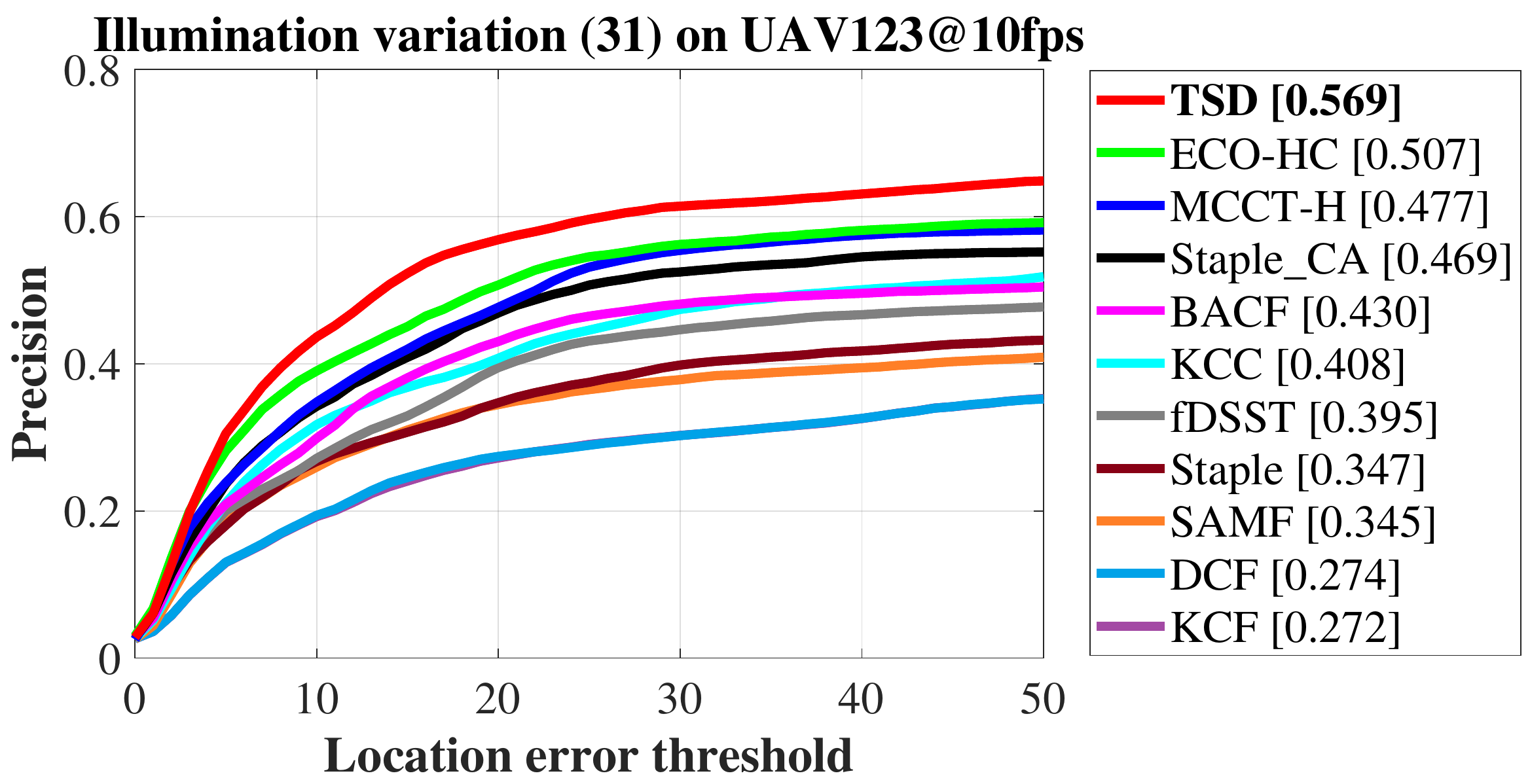}
		\end{minipage}%
	}%
	\subfigure{
		\begin{minipage}[t]{0.32\textwidth}
			\centering
			\includegraphics[width=1\textwidth]{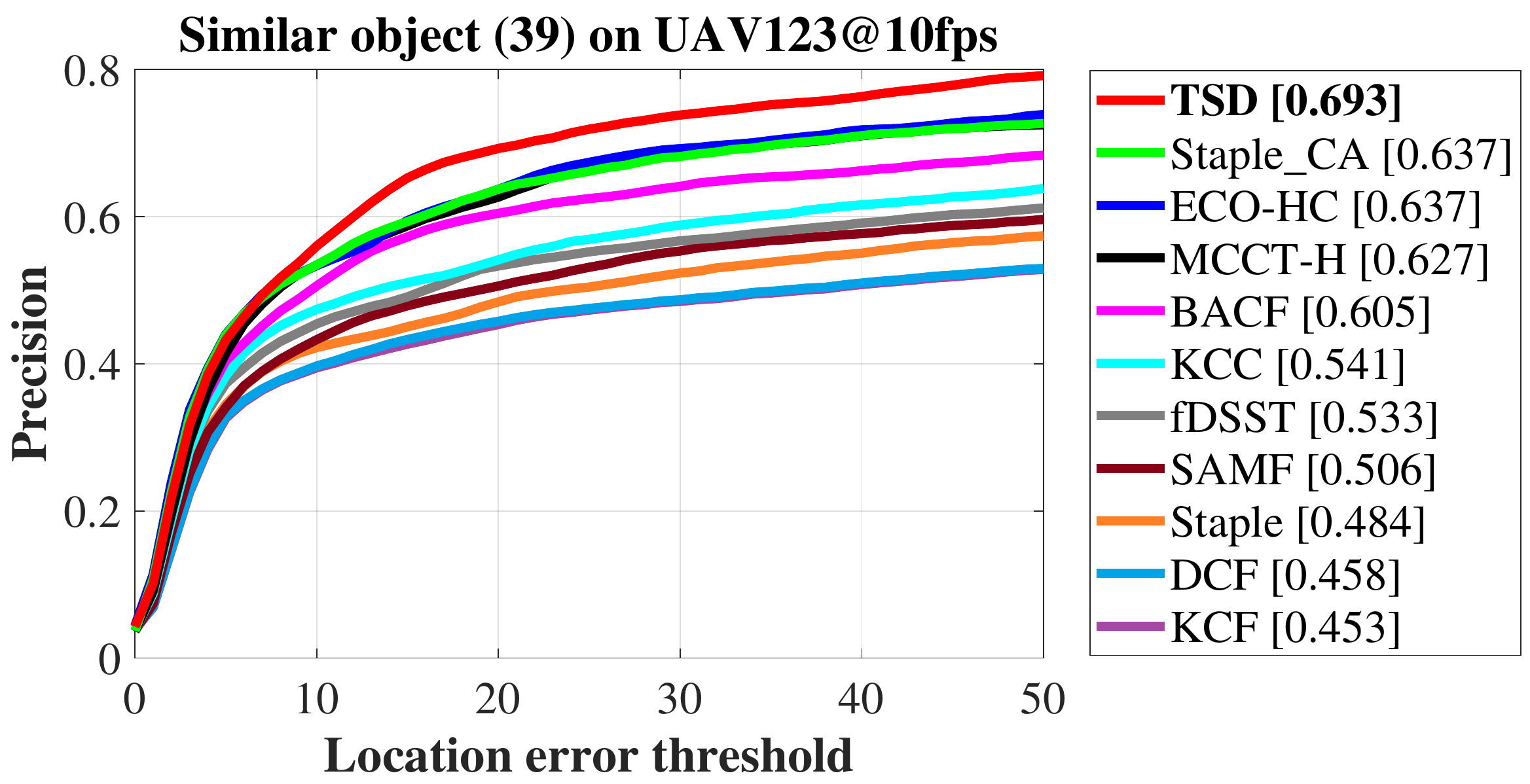}
		\end{minipage}%
	}%
	\centering
	\caption{Precision and success plots of TSD as well as other real-time  tracking approaches on CPU in nine challenging attributes on UAV123@10fps \cite{Mueller2016ECCV}.}
	\label{fig:attribute}
\end{figure*}

\subsubsection{\textbf{Overall performance evaluation}}
The experimental results achieved by TSD tracker and other state-of-the-art trackers on UAV123@10fps \cite{Mueller2016ECCV} and DTB70 \cite{LiAAAI2017} are demonstrated in 
Fig.~\ref{fig:zongtu}, in which precision plots (PPs) and success plots (SPs) are employed for evaluation. 
As shown in the precision plots, TSD outperforms the second-best tracker ECO-HC and the third-best tracker MCCT-H by $3.9\%$ and $10.6\%$, respectively. 
Similar to PPs, TSD is also ranking No.1 among other state-of-the-art trackers in SPs. 
Ranking map on UAV123@10fps in terms of PPs and SPs is displayed in Fig.~\ref{fig:paiming}. 
As for DTB70, TSD outperforms the second-best tracker ECO-HC and third-best tracker MCCT-H by a gain of $2.2\%$ and $8.8\%$ in the precision, respectively. In success plots, TSD is ranking No.2 among 11 state-of-the-art trackers. Yet it is noted that the best tracker ECO-HC \cite{Danelljan2017CVPR} employs both histogram of gradient (HOG) \cite{Henriques2014TPAMI} and CN. 
Besides satisfactory tracking results, the speed of TSD is adequate for real-time UAV tracking applications, as shown in Table~\ref{table:table1}.

\subsubsection{\textbf{Attribute based comparison}}
Besides overall performance, the attribute-based performance of TSD and other trackers are also evaluated. Precision plots in the scenarios of partial occlusion, viewpoint change, out of view, aspect ratio change, camera motion, fast motion, full occlusion, illumination variation as well as similar object around are demonstrated in Fig.~\ref{fig:attribute}. TSD has exhibited a huge improvement from its baseline BACF, and has achieved state-of-the-art performance in all the challenging attributes. In these scenarios, unreliable samples can be easily introduced into training-set. 
Typically, CF-based trackers ignore to manage the training-set. 
TSD is able to improve the quality of the training-set and reduce the unexpected effects caused by unreliable samples efficiently.
\subsection{Ablation study}
The ablation study is conducted on UAV123@10fps \cite{Mueller2016ECCV}. 
Significant performance improvement is achieved by integrating the unreliable sample discarding with the baseline. 
As shown in Table~\ref{table:table2}, it outperforms the baseline in precision plots and success plots by a gain of $11.0\%$ and $7.7\%$, respectively. 
The weighted sample fusion based on the time slot establishment further improves the performance by reducing the influence of untrustworthy samples. 
Additionally incorporating the proposed response map based regularization has elevated TSD to 0.659 and 0.464 in precision and success rate respectively, leading to a final improvement of $15.2\%$ and $12.3\%$ respectively compared to the baseline.
\begin{table}[t]
	\scriptsize
	\setlength{\tabcolsep}{0.5mm}
	\centering
	\caption{Ablation study on UAV123@10fps \cite{Mueller2016ECCV}. Rel. imp. indicates relative improvement compared to the last step.}
	\begin{tabular}{ccccccccccccc}
		\hline\hline
		&\textbf{Baseline}&$\Rightarrow$& \textbf{Unreliable Sample }&$\Rightarrow$&\textbf{Sample}&$\Rightarrow$&\textbf{Response Map}\\
		&\textbf{BACF}&&\textbf{Discarding}&&\textbf{Fusion}&&\textbf{Based Regularization}\\\hline
		\textbf{PPs}&0.572&&0.635&&0.645&&\textbf{0.659}\\
		\textbf{Rel. imp.}&-&&$11.0\%$&&$1.6\%$&&\textbf{$2.2\%$}\\
		\textbf{SPs}&0.413&&0.445&&0.454&&\textbf{0.464}\\
		\textbf{Rel. imp.}&-&&$7.7\%$&&$2.0\%$&&\textbf{$2.2\%$}\\\hline\hline
	\end{tabular}%
	\label{table:table2}%
\end{table}%
\section{Conclusions}
In this work, a novel correlation filter with training-set distillation is proposed for UAV tracking. 
In the process of training-set distillation, historical samples are scored dynamically for enhancing the tracking reliability. 
For efficiency reason, the proposed tracker has employed keyframes to divide the tracking process into multiple time slots. 
In the current time slot, the most unreliable sample will be discarded when the number of current training samples exceeds the given size. 
Samples in the newly established time slot are fused into one sample immediately to decrease the training redundancy. 
Extensive tests have validated that our tracker outperforms significantly better than many state-of-the-art works. 
We believe that our method can further improve the development of UAV tracking.
\section*{Acknowledgment}
The work was supported by the National Natural Science Foundation of China (No.61806148) and the Fundamental Research Funds for the Central Universities (No.22120180009).
%
%
\newpage
\bibliographystyle{IEEEtran}  
\bibliography{egbib}

\end{document}